\documentclass{article}

\usepackage[preprint]{neurips_2026}

\usepackage[utf8]{inputenc} 
\usepackage[T1]{fontenc}    
\usepackage{hyperref}       
\usepackage{url}            
\usepackage{booktabs}       
\usepackage{amsfonts}       
\usepackage{nicefrac}       
\usepackage{microtype}      
\usepackage{xcolor}         
\usepackage{amsmath}
\usepackage{graphicx}
\usepackage{xspace} 
\usepackage{algorithm}
\usepackage{algpseudocode}
\usepackage{listings}
\usepackage{float}
\usepackage{caption}
\usepackage{multirow}
\usepackage{wrapfig}
\usepackage{float}

\usepackage{natbib}
\setcitestyle{authoryear,square,comma}
\let\citet\citep

\definecolor{mydarkblue}{rgb}{0, 0.08, 0.45}
\hypersetup{
  colorlinks=true,
  citecolor=cyan,
  linkcolor=red,
  urlcolor=red
}

\lstdefinestyle{pseudopy}{
    language=Python,
    basicstyle=\ttfamily\small,
    keywordstyle=\bfseries\color{blue!60!black},
    commentstyle=\color{green!70!black},
    stringstyle=\color{purple!60!black},
    showstringspaces=false,
    columns=fullflexible,
    keepspaces=true,
    mathescape=true,
    escapeinside={(*}{*)},
}

\newcommand{\ourmethod}{\emph{TIDE}\xspace}
\newcommand{\minisection}[1]{\vspace{0.05in}\noindent {\bf #1}}

\title{\ourmethod: Efficient and Lossless MoE Diffusion LLM Inference with I/O-aware Expert Offload
}

\author{
Zhiben Chen$^{1,2}$\thanks{Equal Contribution}
\quad Youpeng Zhao$^1$\footnotemark[1] 
\quad Yang Sui$^3$ 
\quad Jun Wang$^1$ 
\quad Yuzhang Shang$^1$\thanks{Corresponding Author} \\[6pt]
$^1$University of Central Florida
\quad $^2$Mobi.AI
\quad $^3$Rice University \\[6pt]
\href{https://tide-paper.vercel.app}{\textbf{Project}}
~~~~~~
\href{https://github.com/ims-kdks/TIDE}{\textbf{Code}}
}

\begin{document}

\maketitle

\begin{abstract}
\label{sec:abstract}
Diffusion Large Language Models (dLLMs) have emerged as a competitive alternative to autoregressive (AR) models, offering better hardware utilization and bidirectional context through parallel block-level decoding. 
However, as dLLMs continue to scale up with mixture-of-experts (MoE) architectures, their deployment on resource-constrained devices remains an open challenge.
Existing AR-based methods often incur either prohibitive I/O overhead or significant compute bottlenecks.
In this work, we propose \textbf{\ourmethod}, a novel resource-efficient inference system that leverages the temporal stability of expert activations during the diffusion process within the block. 
Specifically, we leverage the temporal stability of expert activations during the diffusion process within the block and introduce an interval-based expert refresh strategy that updates the expert placement in an I/O-aware fashion.
To ensure optimal performance, we formulate the inference scheduling as a mathematical programming problem, solving for the optimal interval that minimizes I/O traffic and CPU computation.
Most importantly, \ourmethod is a lossless optimization that requires no model training, providing a "free lunch" acceleration for dLLM inference.
In a single GPU-CPU system, we demonstrate that \ourmethod achieves up to 1.4$\times$ and 1.5$\times$ throughput improvements over prior baselines on \texttt{LLaDA2.0-mini} and \texttt{LLaDA2.0-flash} models, respectively.  
\end{abstract}
\section{Introduction}
\label{sec:intro}

Diffusion-based Large Language Models (dLLMs) have recently emerged as a competitive alternative to autoregressive (AR) Large Language Models (LLMs)~\citep{opt,gpt2,deepseek,mixtral} for text generation tasks. 
Instead of producing tokens one-by-one in a sequentially left-to-right fashion, dLLMs iteratively denoise multiple masked tokens at the granularity of a block, offering two structural advantages over AR models: 
(1) each token prediction is conditioned on bidirectional context, allowing for better semantic understanding, and
(2) multiple tokens within a block can be decoded in parallel to improve computational efficiency. 
Built upon this paradigm, a series of open-sourced dLLMs~\citep{nie2025llada,ye2025dream,cheng2025llada2,fast-dllm,diffusionllama} has emerged, most notably the LLaDA series~\citep{nie2025llada,cheng2025llada2}, which has achieved performance comparable to its AR counterparts while offering much higher decode throughput~\citep{diffusionllama}.
Most recently, LLaDA-2~\citep{cheng2025llada2} adopts a sparse mixture-of-experts (MoE) backbone~\citep{fedus2021switch,deepseek,mixtral} as AR-based models, in which tokens are routed to a small subset of experts at each layer.
This design scales diffusion language models from the original 8B parameters to 100B, making them more production-ready~\citep{dfm1,dfm2}. 

With the ever-increasing popularity of edge computing, running AI models in resource-constrained environments has attracted growing attention in both research and practice~\citep{sheng2023flexgen,mobilellm,merino,alisa}.
Such on-device intelligence both speeds up response latency and enhances data privacy and security, making AI more accessible, efficient, and practical in a wide range of daily applications~\citep{appleai,copilot}. 
Thanks to their inherent parallelism, dLLMs emerged as a compelling option for near-user inference~\citep{fast-dllm,qdllm}.
As the compute capability of edge hardware, such as mobile NPUs and CPUs~\citep{appleai,merino}, continues to scale up, dLLMs have become a much more natural fit for on-device uses, achieving significantly higher hardware utilization than memory-bound operations characteristic of token-by-token AR decoding.

While prior research has achieved promising results on optimizing dense dLLM architectures (typically $<$8B parameters)~\citep{fast-dllm,sun2025dkvcache,l2p,prophet}, 
these methods generally focus on model compression~\citep{q1,q2}, caching~\citep{fast-dllm,sun2025dkvcache}, or efficient decoding~\citep{l2p,prophet}. 
The efficient deployment of Mixture-of-Experts (MoE) dLLMs~\citep{cheng2025llada2} on resource-limited platforms stands as an open question.
Unlike their AR counterparts, MoE-dLLMs present a distinct execution pattern:
\textbf{In MoE-dLLMs, each denoising step activates experts for all tokens simultaneously within the block. 
This produces a wide, fragmented expert footprint that could easily trigger out-of-memory (OOM) errors.
}

A straightforward solution is to swap experts between GPU and CPU memory~\citep{fastio,moeinfinity}.
However, expert migration at every denoising step is prohibitively expensive, as a single dLLM step activates a larger, more diverse set of experts than an AR step, thus creating massive CPU-GPU I/O traffic.
An alternative approach is to simply reroute token computation to the CPU experts~\citep{fiddler}.
But modern CPU execution is often orders of magnitude slower than GPU execution, especially for dense general matrix multiplication (GEMM) operations.
The system inevitably becomes CPU-bound as more tokens are routed to the host, causing the GPU to idle while waiting for CPU-processed activations.
\begin{figure}[t]
    \centering
\includegraphics[width=\linewidth]{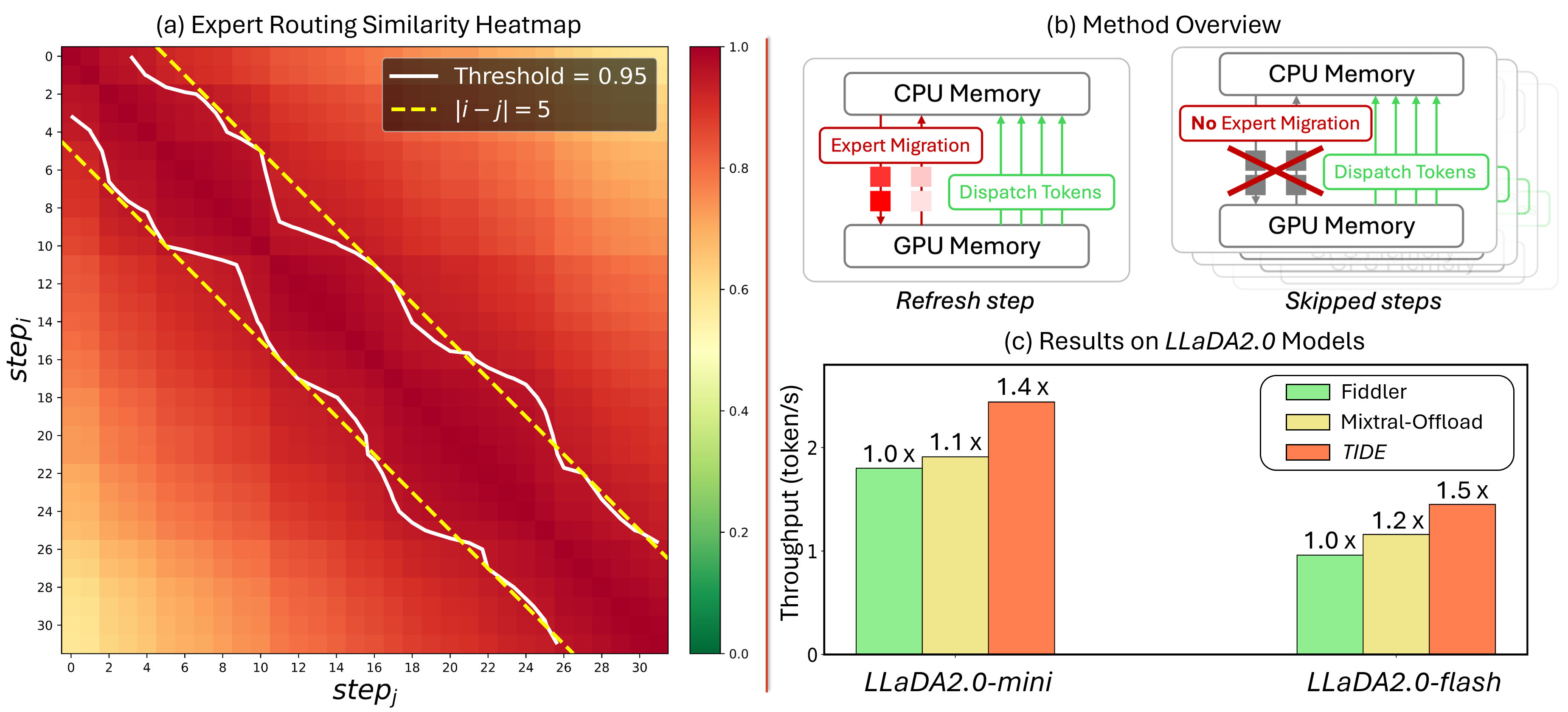}
    \vspace{-18pt}
    \caption{(a)
    Similarity heatmap of expert routing across denoising steps within a block. 
    Expert routing remains highly similar for nearby steps, and the diagonal bands show that this stability extends beyond immediate neighbors: step pairs separated by five denoising steps retain cosine similarity near $0.95$. 
    (b)
    Overview of \ourmethod.
    At \textit{refresh steps}, the system intelligently swaps the GPU and CPU experts based on token hit counts (number of tokens each expert has processed).
    At \textit{skipped steps}, the system continues decoding with the current expert placement and does not migrate experts. 
    By exploiting routing stability across adjacent steps, \ourmethod avoids unnecessary GPU-CPU I/O overhead and maintains high GPU utilization.
    (c)
    Throughput comparison of \ourmethod against state-of-the-art MoE inference solutions~\citep{fiddler,fastio} for \texttt{LLaDA2.0} in a single GPU-CPU setting. 
    }
    \vspace{-0.3in}
    \label{fig:overview}
\end{figure}
Consequently, there is an urgent need for an orchestration strategy that achieves
(1) \textit{minimal I/O overhead} and 
(2) \textit{maximal compute efficiency} in the case of inference on resource-constrained systems.

In this work, we propose \textbf{\ourmethod},
a new I/O-aware MoE-dLLM inference system that intelligently schedules the expert routing decisions to improve system throughput with no accuracy drop.
\textbf{Our key insight is that the expert activation exhibits similar patterns in multiple adjacent denoising steps within a block, thereby creating the opportunity for expert reuse, as shown in Figure~\ref{fig:overview}~(a).}
\ourmethod adopts an interval-based expert refresh and reuses the GPU expert set within the same interval.
\ourmethod aims to maintain a high GPU expert hit rate while reducing expert migration overhead, which is especially costly in dLLMs because each denoising step routes an entire active block rather than a single new token.
Moreover, our method does not require any model training and has no impact on model accuracy, thus offering a free-lunch type acceleration for MoE-dLLM inference.

As shown in Figure~\ref{fig:overview}~(b), \ourmethod splits the decode phase into \textit{refresh steps} and \textit{skip steps}:
At refresh steps, \ourmethod promotes the CPU experts with the most token hits to the GPU memory up to its budget.
For skipped steps, the model reuses the current placement and routes the tokens to their corresponding expert sets in an asynchronous fashion.
The optimal interval is determined by modeling the latency overheads using an analytical model and solving a constrained mathematical programming (MP) problem with a combination of hardware profiling and greedy search.
Evaluations on both \texttt{LLaDA2.0-mini} and \texttt{LLaDA2.0-flash} on NVIDIA A100 and H100 GPUs demonstrate that \ourmethod obtains up to 1.4$\times$ and 1.5$\times$ speedup under different memory constraints against prior works.

In summary, we make the following contributions 
:
\begin{itemize}
    \item We identify the challenges for MoE-dLLM inference and propose a new training-free and lossless solution, \ourmethod, for efficient inference on resource-constrained environments.
    \item By exploiting the cross-step similarity in expert routing, \ourmethod features an interval-based expert refresh strategy that intelligently schedules the expert placement to avoid unnecessary I/O overhead. 
    We optimize the interval choice by formulating and solving the MoE inference as a constrained mathematical programming problem with an analytical model.
    \item We implement and evaluate \ourmethod on \texttt{LLaDA2.0} models in a single GPU-CPU system. 
    Experiments demonstrate that \ourmethod can significantly improve system efficiency over previous baselines without any accuracy drop.
\end{itemize}

\section{Related Work}
\label{sec:related}

\minisection{Diffusion Large Language Models (dLLMs).}
Diffusion models are a class of generative models that learn to transform noise into data through an iterative denoising process~\citep{d1,d2,dit}.
They have been widely adopted in image and video generation, where models start from random noise and progressively refine it into high-quality images or videos that align with a given prompt~\citep{pixart,opensora}.
Recently, combining diffusion models with LLMs has become a promising direction~\citep{nie2025llada,ye2025dream,cheng2025llada2}.
Instead of predicting the very next word, they take a block of random noise—or a sequence of masked tokens—and gradually refine it into coherent text~\citep{nie2025llada,ye2025dream}. 
This decoding structure provides dLLMs with bidirectional context and enables block-level parallelism during generation. 
Recent work shows that such a diffusion-based paradigm can scale to a mixture-of-experts (MoE) architecture, with better improved compute efficiency~\citep{cheng2025llada2}. 
In this work, we focus on improving the inference efficiency for MoE-based dLLMs.

\minisection{Inference Optimization for dLLMs.}
Due to the rising popularity of dLLMs in both academia and industry, there have been several works focusing on improving their inference-time efficiency, especially for dense models~\citep{ye2025dream,nie2025llada}.
A significant body of work focuses on model compression~\citep{q1,q2}, improved caching~\citep{fast-dllm,sun2025dkvcache}, or efficient decoding strategies~\citep{l2p,prophet,apd}.
Notably, following the KV cache mechanism of AR models~\citep{sheng2023flexgen,alisa,vLLM}, Fast-dLLM proposes similar block-wise approximate KV caching and a confidence-aware parallel decoding with minimal quality drop~\citep{fast-dllm}.
dKV-Cache exploits the stable KV states in neighboring states to reduce repeated attention computation~\citep{sun2025dkvcache}.
Learn2PD further introduces a learning-based filter model to avoid redundant decoding and achieve better inference efficiency~\citep{l2p}.
However, from the best of our limited knowledge, there has been no prior work on improving the runtime efficiency of MoE-dLLMs.

\minisection{Mixture-of-Expert (MoE).}
MoE-based models have shown promising performance in a wide range of applications and have become the de facto model choice for real-world production systems~\citep{deepseek,mixtral,rajbhandari2022deepspeedmoe}.
Unlike dense models, MoE architectures increase parameter capacity by increasing the number of FFNs (experts), with a subset of experts activated per token to reduce effective computation relative to the total model size~\citep{fedus2021switch,deepseek,mixtral}.
However, deploying MoE models efficiently is particularly challenging due to their large memory footprint, particularly for resource-constrained scenarios.
Modern GPU memory cannot hold all the expert weights, creating 
additional latency overhead of frequent expert swapping between GPU HBM and CPU host memory~\citep{fastio,moeinfinity} or slow CPU-based computation~\citep{fiddler}.
To make matters worse, a much larger pool of experts is activated at each step due to its parallel processing nature during the dLLM-MoE inference process, thus making prior solutions ill-suited for diffusion-based models.

\section{Methodology}
\label{sec:method}
\begin{figure}[t]
    \centering
    \includegraphics[width=\linewidth]{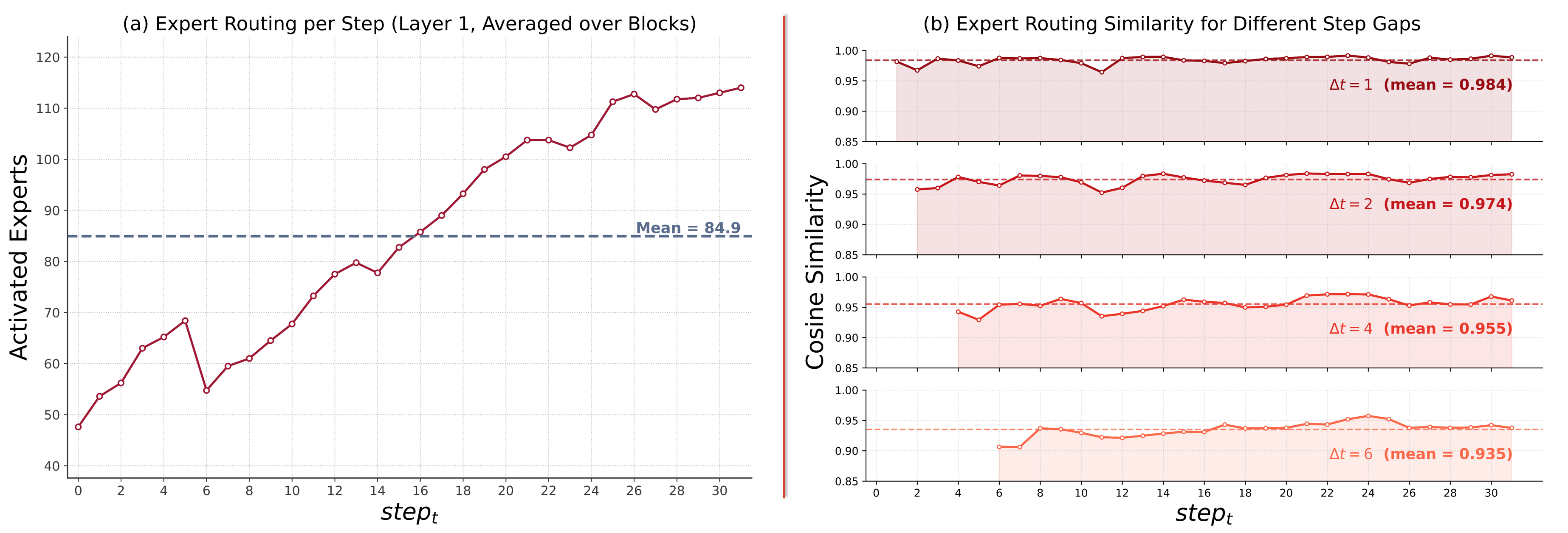}
    \vspace{-18pt}
    \caption{
    Expert activation pattern with a block of size 32 for \texttt{LLaDA2.0-mini}.
    (a) Number of unique experts activated at each step, which increases as decoding continues.
    (b) Similarity scores of expert routing for different step intervals at each step within a block. 
    Here, we use the cosine similarity score, following prior works~\cite{fast-dllm,l2p}.
    }
    \vspace{-0.2in}
    \label{fig:pattern}
\end{figure}
In this section, we begin by introducing some preliminary details on the mixture-of-experts (MoE) 
and formulating the efficiency problem for resource-constrained inference.
Next, we present our observations for the expert activation pattern and key insights.
Finally, we elaborate on our scheduling and execution strategy for MoE-dLLM, which includes 
(1) a mathematical programming (MP) model to determine the optimal interval and 
(2) a detailed description of the expert placement procedure for MoE-dLLM inference.

\subsection{Problem Definition}
\label{subsection:problem}
Consider a MoE model~\citep{deepseek,mixtral}
with $L$ sparse feed-forward network (FFN) 
layers with $E$ total experts, and $k$ experts are activated for one token at a time.
For a batch of tokens, assume $K$ GPU experts are used for token computation, where $K>>k$, the latency of processing tokens in one FFN layer can be formulated as the total of GPU computation time:
\begin{align}
    \mathbf{Lat}^{\text{FFN}} = \mathbf{Lat}^{\text{GPU}}(K)
    \label{eq:1}
\end{align}
In resource-constrained platforms, GPU memory cannot hold all the expert weights for large MoE models. 
For instance, Mixtral-8x7B consists of over 46B parameters, requiring over 94\,GB of GPU VRAM in FP16, exceeding a single H100 80\,GB GPU~\citep{mixtral}.
Here, we assume the GPU holds $B$ number of experts, and the remaining $(E-B)$ experts reside in host memory. 
The expert selections can be divided as $K = \{K^{\text{GPU}}, K^{\text{CPU}}\}$,  and prior methods employ two strategies:
(1) reroute tokens to host memory experts for CPU computation or
(2) swap the experts between the GPU and the host memory.
For token routing, the latency per FFN layer is as follows:
\begin{align}
\mathbf{Lat}^{\text{FFN}} = \text{max}(\mathbf{Lat}^{\text{GPU}}(K^{\text{GPU}}), \mathbf{Lat}^{\text{CPU}}(K^{\text{CPU}})) 
\label{eq:2}
\end{align}
And in the case of expert swapping, the latency can be formulated as the GPU computation time and additional expert I/O transfer latency:
\begin{equation}
\mathbf{Lat}^{\text{FFN}} = \begin{cases} 
      \mathbf{Lat}^{\text{GPU}}(K) + \mathbf{Lat}^{\text{I/O}}(K^{\text{CPU}})& \text{if } K < B \\
      \text{max}(\mathbf{Lat}^{\text{GPU}}(K^{\text{GPU}}), \mathbf{Lat}^{\text{CPU}}(K-B)) + \mathbf{Lat}^{\text{I/O}}(B-K^{\text{GPU}}) & \text{if } K > B 
   \end{cases}
   \label{eq:3}
\end{equation}

We can see that latency is highly dependent on the number of existing GPU experts used for token computation, i.e., the \textit{GPU expert hit rate}.
For single-batch AR decoding, the number of activated experts stays fixed at $K=k$, which presents not much of an obstacle.
However, as shown in Figure~\ref{fig:pattern}~(a), in the case of diffusion-based MoE, experts for all tokens within a block are activated, leading to potentially high $K$.
According to equation~\ref{eq:3}, when the experts on the GPU are not selected for FFN computation $(K>B)$, the inference runtime is bottlenecked by both the CPU computation and GPU-CPU I/O, creating potentially severe inference bottlenecks.
\textbf{Given this efficiency obstacle in MoE inference, we need to find a scheduling policy that orchestrates both the expert migration and token routing in resource-constrained systems, so that the overall execution time is minimized}.

\begin{figure}[t]
    \centering
\includegraphics[width=\textwidth]{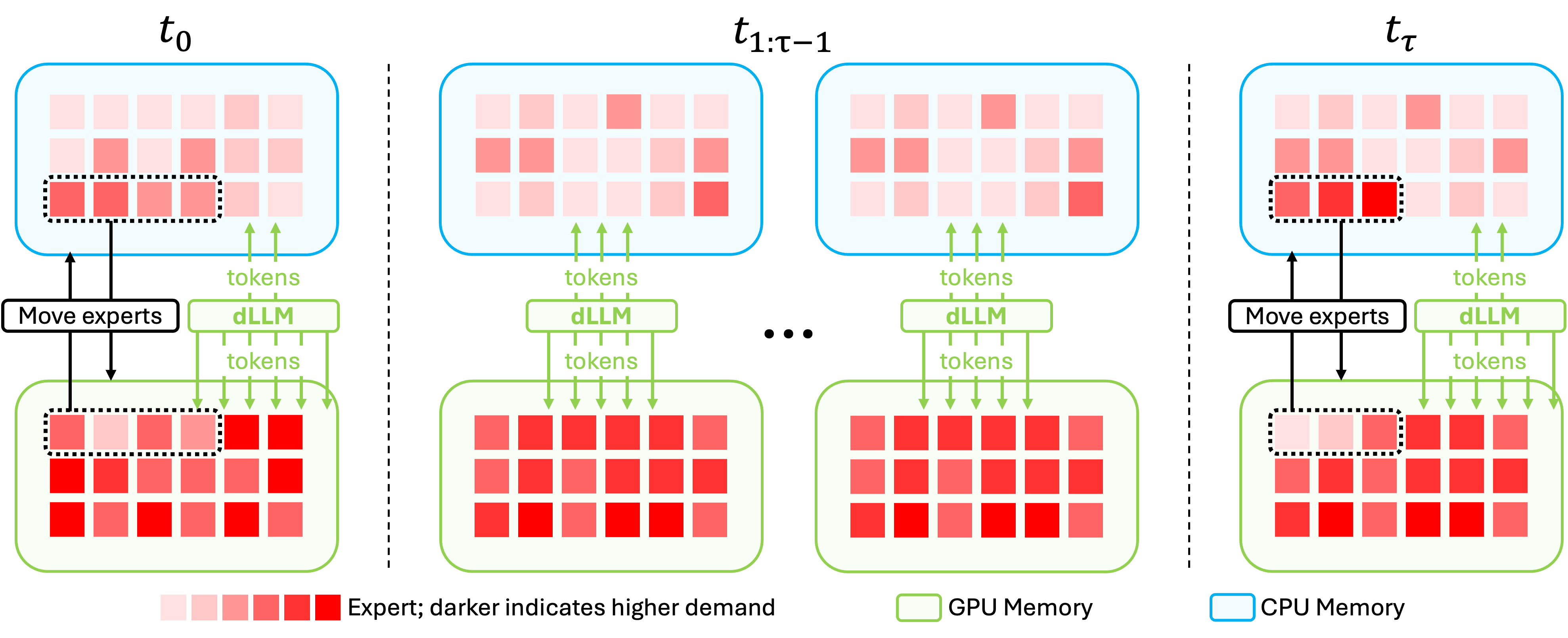}
\vspace{-18pt}
    \caption{Design of \ourmethod. 
    At \textbf{refresh steps} ($t_0, t_{\tau}$), the system updates the GPU-resident expert set by promoting experts with the highest token hits from CPU memory to GPU memory. 
    Experts outside this set are kept in CPU memory or evicted there if currently GPU-resident. 
    At \textbf{skipped steps} ($t_{1:\tau - 1}$), decoding continues with the current expert placement and performs no expert migration. 
    }
    \vspace{-0.2in}
    \label{fig:method_overview}
\end{figure}

\subsection{Observation \& Insights}
Since diffusion-based generation is inherently an iterative reverse process, each token is progressively denoised in a coarse-to-fine manner, evolving from \texttt{[MASK]} toward a concrete \texttt{[word]} prediction~\citep{d1,d2,nie2025llada}. 
As a result, the latent representations produced at adjacent denoising steps often exhibit strong similarity, a property that has been observed in prior work~\citep{sun2025dkvcache,fast-dllm}. 
Motivated by this observation, we investigate whether a similar form of cross-step stability also arises in expert activation during MoE-dLLM inference.

Figure~\ref{fig:pattern}~(b) shows that adjacent denoising steps indeed induce highly similar expert activation patterns. 
We highlight two key observations. 
\underline{First}, expert activation exhibits strong temporal locality. 
The set of activated experts changes only marginally between consecutive denoising steps, with a mean within-block cosine similarity of 0.985. 
This finding aligns with prior observations that intermediate features in diffusion models can be effectively reused across nearby denoising steps~\citep{sun2025dkvcache,fast-dllm,l2p}. 
\underline{Second}, routing similarity remains high not only between immediate neighbors but also within a broader band around the diagonal. 
In particular, step pairs separated by as many as five denoising iterations still retain a cosine similarity above 0.95. 

The above observations indicate that routing decisions at one step are highly predictive of expert demand in subsequent steps, suggesting that the expert activation distribution can be treated as approximately quasi-static over short denoising intervals.
This temporal stability has an important practical implication: rather than recomputing or adapting expert-related decisions independently at every denoising step, MoE-dLLM inference can potentially amortize such decisions across a short window of steps. 
There, it creates an opportunity to exploit routing locality for more efficient inference while preserving the model’s dynamic expert selection behavior.

\subsection{\ourmethod Design}
Given the above-mentioned findings, we propose \ourmethod, which leverages the temporal locality of expert activation patterns to intelligently make the expert swapping and token routing decisions in a training-free manner.
Specifically, \ourmethod introduces an expert refresh strategy that swaps the experts between GPU memory and host memory at the interval of $\tau$ steps $(\tau>1)$ within a block.
As shown in Figure~\ref{fig:method_overview}, \ourmethod partitions the decoding process within a block into two distinct phases: \textit{refresh steps} and \textit{skipped steps}. 
At refresh steps (e.g., $t_0$ or $t_{\tau}$), \ourmethod dynamically updates the GPU-resident expert set by promoting `high-demand' experts on the host with the highest token hits to GPU memory, while evicting `low-demand' experts back to CPU memory. 
During the intervening skipped steps ($t_1$ to $t_{\tau-1}$), a fixed expert placement is maintained with no migration while dispatching tokens to their selected experts. 
This hybrid approach ensures that the majority of computation remains on the GPU, significantly amortizing the overhead of expert swapping and CPU computation.
Since \ourmethod only changes to perform load balancing between GPU and CPU experts, it has no impact on model outputs.

\begin{figure}[t]
    \centering
    \includegraphics[width=\textwidth]{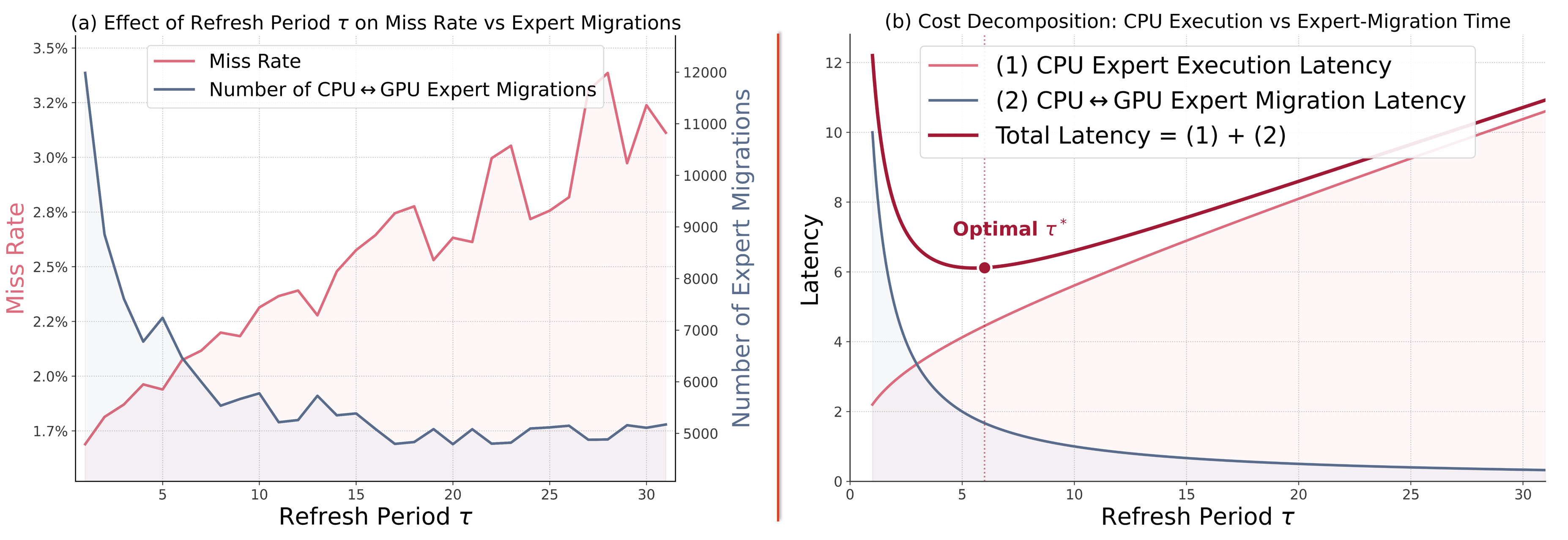}
    \vspace{-18pt}
    \caption{
    Impact of the refresh interval $\tau$. 
    (a) Relationship of GPU expert miss rate and the number of expert migrations with respect to $\tau$.
    Increasing $\tau$ generally raises the GPU expert \textit{miss rate}.
    Meanwhile, a larger $\tau$ reduces the number of expert migrations. 
    The migration curve shows diminishing returns at larger $\tau$, consistent with our drift analysis in Eq.~\ref{eq:migration_count}. 
    (b) Relationship of expert migration and CPU computation latency with respect to $\tau$ based on our analytical model.
    We can see that an optimal $\tau$ can be determined by solving the optimization problem in Eq.~\ref{eq:MP}.
    }
    \vspace{-0.2in}
    \label{fig:tradeoff_analysis}
\end{figure}

\minisection{Interval-based Expert Refresh.}
\label{minisec:analytical_model}
In the design of \ourmethod, a key question to answer is how to decide the optimal interval $\tau$.
As shown in equations~\ref{eq:2} and~\ref{eq:3}, the dominant costs of offloaded MoE inference are expert migration (I/O) and CPU expert executions.
To better understand the tradeoffs between I/O and CPU computation, we define the mismatch between the expert set in step $t-1$ and $t$, as \textit{drift rate} as:
\begin{equation}
    d_{t} = \frac{|\Delta K^{\text{GPU}}_{t}|}{B}
    \label{eq:drift}
\end{equation}
where $\Delta K_{t}^{\text{GPU}}$ denotes the expert selection difference between the current optimal placement and the previous optimal placement.
Assuming independent per-expert replacement events, the probability that any given expert is still optimal after $\tau$ steps is $\prod_{t=1}^{\tau-1}(1-d_{t})$, so the expected number of experts that need to be migrated at the next refresh is $B \cdot (1 - \prod_{t=1}^{\tau-1}(1-d_{t}))$. 
From Figure~\ref{fig:overview}(a), we see that cross-step similarity scores exhibit high consistency, $d_t$ can be approximated as a constant $d$.
Over $T$ denoising steps, the total expected migration latency costs between GPU and CPU can be treated as a function of $\tau$:
\begin{equation}
    \mathbf{Lat}^{\text{I/O}}(\tau) \approx C^{\text{I/O}} \cdot \frac{B \cdot T}{\tau} \cdot \bigl(1 - (1-d)^{\tau}\bigr).
    \label{eq:migration_count}
\end{equation}
where $C^{\text{I/O}}$ is a GPU-CPU I/O bandwidth-related constant. 
At $\tau = 1$ this recovers the full-refresh baseline $M(1) = B \cdot T \cdot d$. As $\tau$ grows, $1 - (1-d)^{\tau} \to 1$ and $M(\tau) \to B T / \tau$, exhibiting the $1/\tau$ scaling and diminishing returns visible in Figure~\ref{fig:tradeoff_analysis}~(a).

However, increasing the refresh interval $\tau$ comes at a cost.
Although adjacent steps have highly similar routing, the similarity drops when steps get farther away, as shown in Figure~\ref{fig:overview}~(a) and Figure~. 
At a refresh step, ${K}^{\text{GPU}}_t$ is set to the current top-$B$ experts to maximize GPU expert hit rate. 
During skipped steps, the token hit rate continues to fall, as shown in Figure~\ref{fig:tradeoff_analysis}~(a), leading to increased CPU computation time.
The expected CPU computation costs can be defined as:
\begin{equation}
    \mathbf{Lat}^{\text{CPU}}(\tau) \approx C^{\text{CPU}} \cdot T \cdot B \cdot f(\tau).
    \label{eq:migration_count}
\end{equation}
where $C^{\text{CPU}}$ is a CPU-related computation constant and $f(\cdot)$ is a general monotonically increasing function for $\tau$, i.e., $\tau_1 \leq \tau_2 \implies f(\tau_1) \leq f(\tau_2)$. 
To find the optimal $\tau$, we need to minimize the total costs, $\mathbf{Lat}^{\text{total}}(\tau)=\mathbf{Lat}^{\text{CPU}}(\tau) + \mathbf{Lat}^{\text{I/O}}(\tau)$, and solve the following mathematical programming (MP) problem:
\begin{equation}
\label{eq:MP}
    \min_{\tau \in [1,2, ..., T-1]}  (\frac{B \cdot T}{\tau} \cdot \bigl(1 - (1-d)^{\tau}\bigr) + C^{\text{CPU}} \cdot T \cdot B \cdot f(\tau)
\end{equation}
We solve this problem by first running a hardware profiling on the CPU computation speed and I/O bandwidth performance to approximate constants $C$ with different prompt and output configurations to create a mapping between input configurations and their execution time.
Next, we apply a greedy search method to solve the optimization problem for the best performance.
This process is done offline, introducing no overhead during the actual inference process.

\minisection{Expert Selection and Token Routing.}
Another key question to answer is how to perform appropriate expert swapping, i.e., which experts are offloaded and uploaded.
To this end, we employ a global hit counter, where we calculate the expert activation hits for all the experts at refresh steps, select the top experts by frequency ranking, and swap the expert sets on GPU and CPU to maximize reuse potential during skipped steps.
To further minimize the latency overhead of remaining CPU computations, we implement an asynchronous execution pipeline. 
When a token is routed to a host-resident expert (a "miss"), the GPU does not stall. 
Instead, the token features are offloaded to the CPU for concurrent processing while the GPU continues to execute the "hits" for other tokens in the batch. 
The results are re-synchronized at the end of the FFN block, effectively overlapping the slower CPU computation with the high-throughput GPU execution.
The details of our scheduling policy during MoE-dLLM inference are shown in Algorithm~\ref{alg:1}.

\minisection{Lossless Inference.}
Since \ourmethod focuses on expert placement without modifying the selection of MoE router or model weights, where each token is assigned to the same set of experts as GPU-only execution, and does not alter the parallel decoding mechanism, it preserves the model outputs, thus inducing no accuracy degradation.
Our method is essentially lossless, offering free-lunch style inference improvement for MoE-dLLM on resource-constrained platforms.
\begin{algorithm}[!t]
\begin{algorithmic}[1]
\Require Full expert set $\mathcal{E}$ with CPU expert set $\mathcal{E}^{\small\text{CPU}}$ and GPU expert set $\mathcal{E}^{\small\text{GPU}}$, the number of GPU experts $B$, Hit counter $\mathcal{H}$, refresh interval $\tau$, block size $T$ and decoding step $t$, token states $X=\{(x, \mathcal{E}_{x})\}$ has its token $x$ and corresponding expert selections $\mathcal{E}_{x}$.
\For{ all $t<T$}
\If{$t\,\%\,\tau == 0$} \Comment{Update Expert Placement at $\tau$ intervals}
\State $\mathcal{E}_{B}^{\small\text{CPU}},\mathcal{E}_{B}^{\small\text{GPU}} = \text{argmax}_{B} {\mathcal{H}^{\small\text{CPU}}}, \text{argmin}_{B} {\mathcal{H}^{\small\text{GPU}}}$ \Comment{Get the experts to be migrated}
\State $\mathcal{E}_{B}^{\small\text{GPU}} \xrightarrow[\text{Async}]{\text{PCIe}} \mathcal{E}^{\small\text{CPU}}, \mathcal{E}_{B}^{\small\text{CPU}} 
\xrightarrow[\text{Async}]{\text{PCIe}} \mathcal{E}^{\small\text{GPU}}$ \Comment{Asynchronous migrate experts}
\EndIf
\State \textcolor{gray}{\# Token Routing}
\For{ all $(x, \mathcal{E}_{x})$ in $X$} \Comment{Token Routing to Their Respective Experts}
\For{ all $e$ in $\mathcal{E}_{x}$}
\If{$e$ in $\mathcal{E}^{\small\text{CPU}}$}
\State $ x \xrightarrow[\text{Async}]{\text{PCIe}} \mathcal{E}^{\small\text{CPU}}$ \Comment{Route tokens to CPU experts}
\EndIf
\State $ \text{output} = e(x)$ \Comment{Asynchronously Process Tokens on both GPU and CPU}
\EndFor
\EndFor
\EndFor
\end{algorithmic}
\caption{\ourmethod Scheduling Policy for dLLM-MoE Inference}
\label{alg:1}
\end{algorithm}

\section{Experiments}
\renewcommand{\arraystretch}{1.05}
\begin{table}[!t]
\centering
\caption{Performance comparison of our method with Fiddler~\citep{fiddler} and Mixtral-Offload~\citep{fastio} for \texttt{LLaDA2.0} models on the sanitized MBPP benchmark. 
TPS denotes the number of tokens decoded per second (higher is better).
All runs use a block length of 32 and a confidence threshold of 0.95. 
}
\label{tab:main_results}
\resizebox{\columnwidth}{!}{
\begin{tabular}{ccccccc}
\toprule
\multirow{2}{*}{}  & \multirow{2}{*}{\begin{tabular}[c]{@{}c@{}}Gen \\ Length\end{tabular}} & \multirow{2}{*}{\begin{tabular}[c]{@{}c@{}}GPU Expert \\ Budget\end{tabular}} & \multirow{2}{*}{\begin{tabular}[c]{@{}c@{}}GPU Memory\\ Constraint\end{tabular}} & \multicolumn{3}{c}{Throughput (token/s) $\uparrow$} \\ \cmidrule{5-7} 
 &  &   & & Fiddler & Mixtral-Offload & \ourmethod (Ours) \\ \midrule
\multirow{4}{*}{\texttt{LLaDA2.0-mini}}  & \multirow{2}{*}{256} & 64 & 10\,GB & 1.81 &  1.69 &  \textbf{2.11}   \\ 
&  & 128 & 18\,GB & 1.74 & 1.76  &  \textbf{2.36}    \\
 & \multirow{2}{*}{1024}  & 64 & 10\,GB & 1.79  & 1.45 &  \textbf{1.89}  \\
 & & 128 & 18\,GB & 1.80 & 1.91  &  \textbf{2.44} \\ \midrule 
\multirow{4}{*}{\texttt{LLaDA2.0-flash}}  & \multirow{2}{*}{256}  & 32 & 30\,GB & 0.95 &  1.01 &  \textbf{1.25}    \\
 & & 64 & 55\,GB & 1.14  &  1.35 &  \textbf{1.73}    \\
 & \multirow{2}{*}{1024} & 32 & 30\,GB & 0.89  & 1.01  & \textbf{1.24}   \\
 & & 64 & 55\,GB & 1.05 & 1.16  &  \textbf{1.45}  \\ \bottomrule
\end{tabular}
}
\vspace{-0.2in}
\end{table}

\label{sec:experiments}
\subsection{Experimental Settings}
\minisection{Models and Datasets.} 
We evaluate our method on the \texttt{LLaDA2.0} architecture~\citep{cheng2025llada2}, namely, \texttt{LLaDA2.0-mini} (16BA1B) and \texttt{LLaDA2.0-flash} (100BA6B).
Both models have a total of 256 FFN experts with $top\_k=8$ activation pattern.
We use the sanitized MBPP dataset~\citep{austin2021program} from the \texttt{lm\_eval\_harness} library~\citep{eval-harness}.
Since our method is essentially lossless, it can be easily generalized to other datasets.

\minisection{Hardware and Implementations.} 
We run \texttt{LLaDA2.0-mini} on an NVIDIA A100 40\,GB GPU and \texttt{LLaDA2.0-flash} on an NVIDIA H100 80\,GB GPU, with a 48-Core Intel CPU and 1024\,GB DDR4 host memory. 
To ensure high portability and reproducibility, we implement \ourmethod on top of HuggingFace Transformers~\citep{huggingface} and dInfer~\citep{dinfer}, with PyTorch 2.9, CUDA 12.8, which can be further incorporated into popular serving frameworks, such as SGLang~\citep{sglang} and vLLM~\citep{vLLM}.

\minisection{Baselines and Metrics.} Since there is no prior work on optimizing dLLM-MoE inference, we compare \ourmethod with two prior baseline methods for AR-MoE models. 
Specifically, we use Fiddler~\citep{fiddler} as a CPU computation baseline, where expert placements remain static during decoding, and Mixtral-Offloading~\citep{fastio}, which performs expert offloading at each denoising step.
We compare \ourmethod and prior baselines by evaluating inference efficiency with decode throughput, calculated as the average number of decoded tokens per second (token/s).

\subsection{Main Results}

Table~\ref{tab:main_results} demonstrates the system efficiency comparison of our methods and prior solutions.
Overall, we observe that \ourmethod offers the highest attainable throughput for MoE-dLLM inference in resource-constrained systems.
There are three key observations.
First, \ourmethod achieves consistent speedup over all baselines, showing 1.2$\sim$1.4$\times$ higher throughput over Mixtral-Offload.
This is due to the fact that prior works do not consider the unique expert activation pattern for diffusion-based MoE models, thereby leading to suboptimal performance.
Second, the speedup of \ourmethod is consistent across different generation lengths.
While Fiddler achieves comparable performance under certain settings, its efficacy drops when scaling to the larger 100B model with more intensive computation.
Third, \ourmethod shows much better performance, especially when GPU memory capacity is limited.
This is due to the intelligence expert migration policy of \ourmethod, which maximizes the expert reuse and avoids redundant I/O transfer overheads.

\subsection{Performance Analysis}
\minisection{Impact of the Refresh Interval $\tau$.} 
Here, we showcase the efficacy of our refresh interval optimization method.
We compare the throughput performance with different refresh interval configurations, i.e., (1) $\tau=1$ (Mixtral-Offload~\citep{fastio}), (2) random choice, and (3) optimized $\tau$.
We conduct experiments on both \texttt{LLaDA2.0-mini} and \texttt{LLaDA2.0-flash} with varying GPU expert budgets (64/128), varying block lengths (32/64) with confidence threshold 0.95, and generation length 1024, with results shown in Table~\ref{tab:ab_1}.
Here, we summarize two insights.
First, our method consistently delivers the highest throughput out of the three interval choices.
Notably, it provides up to 1.4$\times$ speedup against random $\tau$, which validates our approach of formulating and solving the mathematical programming problem.
Second, \ourmethod can sustain robust performance improvement with respect to both block sizes and GPU budget experts.
It can achieve better speedup against prior baseline~\citep{fastio}, especially in the case of higher GPU expert budgets. 
This is thanks to the interval-based strategy that avoids redundant I/O transfer of expert weights.

\begin{figure}[t]
    \centering
\includegraphics[width=\linewidth]{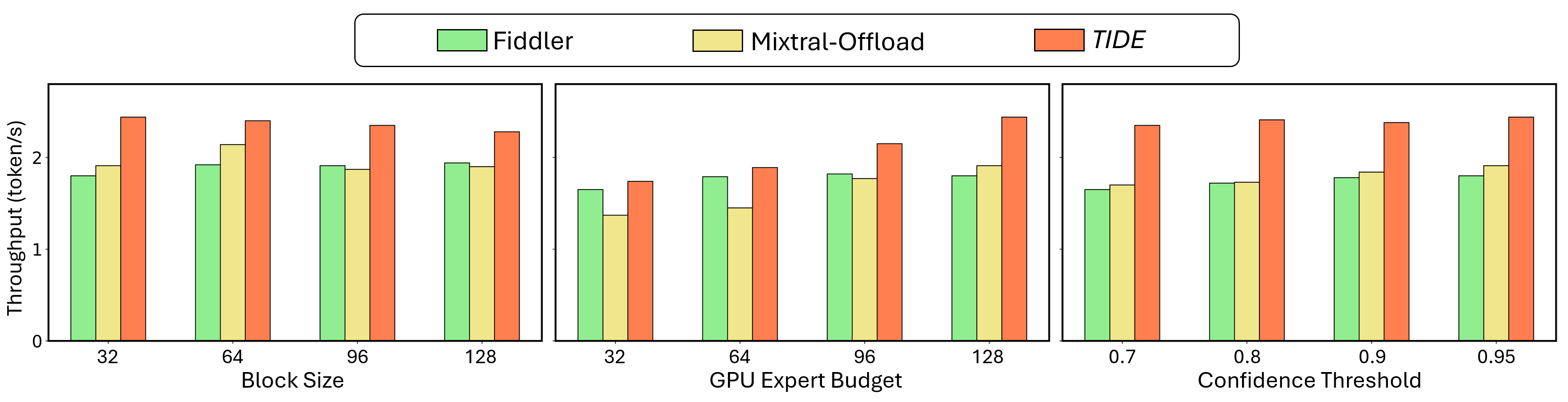}
    \vspace{-16pt}
    \caption{Performance analysis for \texttt{LLaDA2.0-mini} on NVIDIA A100 40\,GB GPU. 
    From left to right are the throughput comparisons of different methods over varying block sizes (32$\sim$128), GPU expert budgets (32$\sim$128), and confidence thresholds (0.7$\sim$0.95).
    We can see that \ourmethod consistently outperforms baseline methods regardless of decoding settings.
    }
    \label{fig:ab}
\vspace{-0.2in}
\end{figure}

\minisection{Sensitivity Studies.}
We further analyze the impact of block size, GPU expert budget, and confidence threshold on the end-to-end system throughput of \texttt{LLaDA2.0-mini} on NVIDIA A100 40\,GB GPU, with results shown in Figure~\ref{fig:ab}.
There are three key observations here.
First, \ourmethod achieves the highest throughput performance against different block sizes with consistent improvement against baseline methods.
This is due to the fact that our interval-based strategy can 
(1) improve GPU expert hit rate, reduce the tokens routed to CPU, thus maximizing compute efficiency, 
and 
(2) minimize I/O expert transfer overhead by using the optimal interval determined by solving an optimization problem.
Second, both Mixtral-offload and \ourmethod scales well with GPU expert budgets, i.e., GPU memory constraints, while Fiddler does not.
This highlights the importance of expert placement in MoE-dLLM inference, since Fiddler is increasingly bottlenecked by CPU computation.
Third, \ourmethod also sustains consistent improvement under different confidence thresholds, with an average speedup of 1.4$\times$ over prior baselines.

\renewcommand{\arraystretch}{1.1}
\begin{table}[!t]
\centering
\caption{Throughput comparison of different interval choices of $\tau$ for \texttt{LLaDA2.0} models on the sanitized MBPP benchmark with varying block sizes and GPU expert budgets.
}
\label{tab:ab_1}
\resizebox{.9\columnwidth}{!}{
\begin{tabular}{cccccc}
\toprule
\multirow{2}{*}{}  & \multirow{2}{*}{\begin{tabular}[c]{@{}c@{}}Block \\ Size\end{tabular}} & \multirow{2}{*}{\begin{tabular}[c]{@{}c@{}}GPU Expert \\ Budget\end{tabular}} & \multicolumn{3}{c}{Throughput (token/s)} \\ \cmidrule{4-6} 
 &  &  & $\tau=1$ & Random $\tau$ & Optimal $\tau$ \\ \midrule
\multirow{4}{*}{\texttt{LLaDA2.0-mini}}  & \multirow{2}{*}{32} & 64 & 1.79  & 1.62 &  \textbf{1.89}   \\ 
&  & 128  & 1.91 & 2.14  & \textbf{2.44}   \\
 & \multirow{2}{*}{64}  & 64  & 1.71  & 1.65 &  \textbf{1.90}  \\
 & & 128  & 2.14 &  2.24 &  \textbf{2.40} \\ \midrule 
\multirow{4}{*}{\texttt{LLaDA2.0-flash}}  & \multirow{2}{*}{32}  & 32  & 1.01  & 0.85  & \textbf{1.24}    \\
 & & 64  & 1.16 & 1.23  &  \textbf{1.45}    \\
 & \multirow{2}{*}{64} & 32  & 1.12  & 0.95 & \textbf{1.32}   \\
 & & 64 & 1.27 & 1.32 &  \textbf{1.49}  \\ \bottomrule
\end{tabular}
}
\vspace{-0.12in}
\end{table}

\section{Conclusion}
\label{sec:conclusion}
This paper proposes \ourmethod, a resource-efficient and I/O-aware inference system for MoE-based diffusion language models.
By exploiting the unique expert activation patterns during decoding, \ourmethod utilizes an interval-based expert refresh strategy to update expert placement periodically, thus avoiding redundant I/O transfer and CPU computation, and the optimal interval is determined by solving an optimization problem, thereby improving end-to-end system decode throughput in resource-constrained scenarios.
Evaluations demonstrate that \ourmethod achieves up to 1.4$\times$, and 1.5$\times$ throughput improvement on \texttt{LLaDA2.0-mini} and \texttt{LLaDA2.0-flash} models, respectively, in a single GPU-CPU system.

\section*{Acknowledgments}
This work was sponsored in part by the Lambda Research Grant and the U.S. National Science Foundation (NSF) under Grants 1907765, 2400014, and 2426368. This work also used Delta at UIUC NCSA through allocation CIS250367 and 250473 from the Advanced Cyberinfrastructure Coordination Ecosystem: Services \& Support (ACCESS) program, which is supported
by U.S. NSF grants 2138259, 2138286, 2138307, 2137603,
and 2138296.

\bibliography{references}

\appendix
\appendix
\section{Appendix: Limitation}
\label{sec:limitations}
As no research is perfect, our work has several limitations as well. 
First, our framework explores the expert activation pattern only within the block, due to its straightforwardness.
Further block-level similarity analysis can provide more insights into the MoE-dLLM decoding procedure, thereby yielding potentially more performance improvement.
Second, the evaluation for this work is performed on limited hardware platforms.
Future work should include explorations of AMD GPUs and ARM CPUs for more comprehensive analysis.
Third, our work is currently limited to resource-constrained settings, e.g., single GPU-CPU systems.
We recognize our insights can be applicable in distributed inference with expert parallelism.
Extension of our work to multi-GPU or even multi-node is an important future avenue of research.
\newpage

\end{document}